\documentclass{article}
\usepackage{amsmath,graphicx,mlspconf}
\usepackage[usenames, dvipsnames]{color}
\usepackage[nocompress]{cite}
\usepackage{srcltx}
\usepackage{amsfonts}
\usepackage[bottom]{footmisc}

\usepackage{amssymb}	
\usepackage{amsfonts}
\usepackage{graphics}
\usepackage{graphicx}
\usepackage{psfrag}
\usepackage{epsfig}
\usepackage{color}
\usepackage{subcaption}
\usepackage{verbatim}
\usepackage{array}
\usepackage{algorithm}
\usepackage{algorithmic}
\usepackage{pifont}
\usepackage{makeidx}  
\usepackage{bbm}
\usepackage{caption}
\usepackage{hhline}

\usepackage{yfonts}
\usepackage{breqn}
\def\BibTeX{{\rm B\kern-.05em{\sc i\kern-.025em b}\kern-.08em
     T\kern-.1667em\lower.7ex\hbox{E}\kern-.125emX}}

\usepackage{color}

\usepackage{url}
\usepackage{dsfont}

\usepackage[square, comma, sort&compress, numbers]{}

\newcommand{\mygamma}{\ensuremath{\gamma{} }}
\newcommand{\mypi}{\ensuremath{\pi{} }}

\usepackage[skip=2pt]{caption}

\copyrightnotice{978-1-7281-0824-7/19/\$31.00 {\copyright}2019 IEEE}

\begin{document}

\toappear{2019 IEEE International Workshop on Machine Learning for Signal Processing, Oct. 13--16, 2019, Pittsburgh, PA, USA }
\title{RL-NCS: Reinforcement Learning Based Data-driven Approach for Nonuniform Compressed Sensing}
\name{Nazmul Karim, Alireza Zaeemzadeh, and Nazanin Rahnavard\thanks{This material is based upon work supported by the National Science Foundation under Grant No. ECCS-1810256 and CCF-1718195.}} 
\address{School of Electrical and Computer Engineering ~\\  University of Central Florida, Orlando, USA ~\\ Emails: nazmul.karim18@knights.ucf.edu, zaeemzadeh@eecs.ucf.edu, nazanin@eecs.ucf.edu}

\maketitle
\begin{abstract}
\small{A reinforcement-learning-based non-uniform compressed sensing (NCS) framework for time-varying signals is introduced. The proposed scheme, referred to as RL-NCS, aims to boost the performance of signal recovery through an optimal and adaptive distribution of sensing energy among two groups of coefficients of the signal, referred to as region of interest (ROI) coefficients and non-ROI coefficients. The coefficients in ROI usually have greater importance and need to be reconstructed with higher accuracy compared to non-ROI coefficients. 
In order to accomplish this task, the ROI is predicted at each time-step using two specific approaches. 
One of these approaches incorporates a long short-term memory (LSTM) network for the prediction. The other approach employs the previous ROI information for predicting the next step ROI.  
Using the \emph{exploration-exploitation} technique, a Q-network learns to choose the best approach for designing the measurement matrix. Furthermore, a joint loss function is introduced for the efficient training of the Q-network as well as the LSTM network. The result indicates a significant performance gain for our proposed method, even for rapidly varying signals and reduced number of measurements.}        
\end{abstract}

\section{Introduction} \vspace*{-\baselineskip} \vspace{0.15cm}  The compressed sensing (CS)~\cite{candes2008introduction, donoho2006compressed} framework aims to recover sparse signals by acquiring significantly fewer number of measurements compared to the classical Nyquist rate. 
Consider a CS problem with the objective of reconstructing a time series $\{\boldsymbol{x}_1,\boldsymbol{x}_2, \ldots ,\boldsymbol{x}_t, \ldots \}$, where $\boldsymbol{x}_t$ is the signal at time step $t$, from an under-sampled time series of linear measurements $\{\boldsymbol{y}_1,\boldsymbol{y}_2, \ldots ,\boldsymbol{y}_t, \ldots \}$, where  $\boldsymbol{y}_t = \boldsymbol{\Phi}_t \boldsymbol{x}_t + \boldsymbol{n}_t$. Here, $\boldsymbol{\Phi}_t \in \mathbb{R}^{M \times N}$ is the measurement matrix and $\boldsymbol{n}_t$ represents the noise that corrupts our measurements at time $t$.
Due to the time-varying nature of our setup, a fixed measurement matrix can deteriorate the reconstruction performance. 
Therefore, our aim is to recover the signal $\boldsymbol{x}_t$ from $\boldsymbol{y}_t$ incorporating a non-uniform sensing strategy attainable by an \emph{adaptive} design of $\boldsymbol{\Phi}_t$. 

In many applications, such as dynamic MRI~\cite{vaswani2008kalman}, high speed video streaming~\cite{asif2010streaming}, and wireless sensor networks~\cite{zaeemzadeh2018cospot}, it is desirable to recover signal that contains coefficients with different levels of importance. For example, a certain area or segment might be more important and informative than the rest of the image. In most cases, the signal coefficients in the $\textit{region of interest}$ (ROI) needs to be reconstructed more accurately. To achieve this, it is required to sample these coefficients with more sensing energy utilizing a non-uniform CS strategy~\cite{Rahnavard2011, Shahrasbi2016HMT}. 
However, if the ROI changes over time, this task gets more complicated and reconstruction using fewer measurements becomes erroneous. 
In this scenario, utilizing the knowledge of previous estimations, we can predict the ROI and design the $\boldsymbol{\Phi}$ accordingly. Hence, a method to make a prediction about the ROI and adaptively design the measurement matrix is introduced in this paper.

We propose a  \emph{reinforcement learning (RL) based adaptive CS technique} that focuses on designing non-uniform measurement matrices for time-varying signals. The objective is to implement a non-uniform sampling method to boost the recovery performance that results in small reconstruction error. The technique of reinforcement learning~\cite{mnih2013playing} has been used in many aspects of wireless sensor networks~\cite{chincoli2018self, le2016reinforcement}. Especially, when a part of the system, an agent, interacts with another dynamic part of the system, known as environment, to come up with the optimal policy that serves the purpose of that system. The decision making ability of an agent based on the past experiences is the key part of an RL. In our work, we have formulated two mechanisms (actions) to predict the next step ROI. At each step, the task of the RL is to choose the best mechanism for deigning the measurement matrix. 


Unlike adaptive CS~\cite{malloy2014near, braun2015info}, the signals that we recover here are not static over time. We also take a different approach than dynamic CS methods~\cite{ziniel2013dynamic,Ziniel2014} that focus only on the reconstruction phase. 
Furthermore, there have been studies that capitalize the idea of utilization of knowledge from prior estimations and they seem to work only if the signal ROI changes very slowly ~\cite{vaswani2010ls, zaeemzadeh2017adaptive}. However, our focus is to obtain adaptive design of CS measurement matrices that show effectiveness even for rapidly varying signals. 
In our work, an agent has the flexibility to choose between two mechanisms compared to only a single mechanism in other works. Furthermore, we propose a multi-task training procedure for tuning two neural networks, the Q-network and the long short-term memory (LSTM)~\cite{greff2016lstm} network. Although a readily available dataset is required to train an LSTM network, we devised an efficient way to tackle this challenge utilizing the experiences stored in the replay memory of the Q-network. In the end, we carried out experiments to prove the superior performing ability of our method in contrast to uniform CS method as well as other techniques.
 
The organization of rest of the paper is as follows. System model and problem formulation have been discussed in Section \ref{sec:SM}. In Section \ref{sec:RL}, we talk about reinforcement learning and Q-learning. A detailed discussion about the approach of our proposed framework has been presented in Section \ref{CSRL}. Then, we describe the experimental setup with simulation results in Section \ref{NE} and conclude with the discussion in Section \ref{Discussion}. 

\vspace*{-\baselineskip} \vspace{0.075cm}
\section{System Model} \label{sec:SM} \vspace*{-\baselineskip} \vspace{0.15cm} 
In compressed sensing, it has been proven that a compressible signal $\boldsymbol{x}_t \in \mathbb{R}^{N}$ is recoverable from relatively fewer random projections, $\boldsymbol{y}_t \in \mathbb{R}^{M}$. 
Consider a scenario where we have a vector-valued time-series that consists of compressible signals $ \{\boldsymbol{x}_1, \boldsymbol{x}_{2},\ldots ,\boldsymbol{x}_t, \ldots \}$. Therefore, the set of measurement vectors can be obtained as
\begin{equation}
    \boldsymbol{y}_t = \boldsymbol{\Phi}_t \boldsymbol{x}_t + \boldsymbol{n}_t , \hspace{0.2cm} \text{for t} \geq 1,
\end{equation} where $\boldsymbol{\Phi}_t \in \mathbb{R}^{M \times N}$ is employed for the mapping of signals to linear measurements at each time-step $t$. The noise term $\boldsymbol{n}_t \in \mathbb{R}^{M} $ is the additive white Gaussian noise (AWGN) with $\boldsymbol{n}_t \sim \mathcal{N}(\boldsymbol{0},\sigma_n^2 \boldsymbol{I}_M)$, where $\boldsymbol{I}_M$ is an all-one vector. It is assumed that all the coefficients of signal $\boldsymbol{x}_t$ are not equally important. 
Therefore, it is necessary to identify and predict the $\textit{region of interest}$ (ROI) that contains the most important coefficients. For this, we aim to predict the ROI instantly based on the knowledge of the signal history. This task of instantaneous decision-making about the position of next ROI can be accomplished with reinforcement learning. After that, it is straightforward to distribute the sensing energy according to the importance levels of the coefficients, such that the coefficients in ROI are being sampled with more sensing energy.    


\vspace*{-\baselineskip} \vspace{0.15cm}
\section{Reinforcement Learning} \label{sec:RL} \vspace*{-\baselineskip} \vspace{0.15cm} In reinforcement learning (RL), an agent tries to learn the optimal policy for a sequential decision-making problem through the optimization of cumulative future reward signal. For solving sequential decision-making problems, it is necessary to estimate the value of each decision or action. 
The action-value function defines how good an action, $a$, is for the agent to take if it is in a state, $\boldsymbol{s}$. In general, a policy $\pi: \boldsymbol{S}\rightarrow \boldsymbol{A}$ dictates the agent which action  it needs to take, from the action space $\boldsymbol{A}$, based on the current state. Given the policy $\pi$, the action value function for a state-action pair $(\boldsymbol{s},a)$ can be determined by 
\begin{equation} \label{bellman}
    Q_{\pi}(\boldsymbol{s},a)= \mathrm{E}_{\pi} \left [\sum_{t=0}^\infty \gamma^t r_t(\boldsymbol{s}_t, a_t)|\boldsymbol{s}_0 = \boldsymbol{s}, a_0 = a, \pi \right].
\end{equation} Here, $r_t$ is the reward at time $t$ and the discount factor $\gamma \in [0,1]$ decides how important the immediate reward is compared to the future rewards. Therefore, $Q_{\pi}$ is the expected sum of future rewards for a state-action pair $(\boldsymbol{s},a)$, also known as Q-function. The optimal policy, $\mypi^*$ is attainable by choosing the action that gives the optimal action-value, $Q^{*}(\boldsymbol{s},a) = \max_\pi Q_{\pi}(\boldsymbol{s},a) $ in each state.

The task of Q-learning~\cite{watkins1992q} can be accomplished by a multi-layered neural network, also known as deep Q-network (DQN). For a given state $\boldsymbol{s}_t$, the output of DQN is a vector of action values $\boldsymbol{Q}(\boldsymbol{s}_t,a;\boldsymbol{\theta}_t)$. 
The two key elements required for tuning the Q-network are (i) \textit{replay memory} which stores observed transitions (experiences)  $(\boldsymbol{s}_t$,$a_t$,$\boldsymbol{s}_{t+1}$,$r_{t+1})$ for a later use, and (ii) a \emph{target network} that generates target, $\beta_t^Q$, for the DQN (online network)~\cite{mnih2015human}. The target network is the same as the online network except with static parameters $\boldsymbol{\theta}^{-}_t$, a periodically copied version of $\boldsymbol{\theta}_t$ (every $\tau$ steps). 
The target for online network stands as
\begin{equation} \label{target}
    \beta_t^Q = r_{t+1} + \mygamma \max\limits_{a \in A} \boldsymbol{Q}(\boldsymbol{s}_{t+1},a; \boldsymbol{\theta}_t^{-}),
\end{equation}
where $Q(\boldsymbol{s}_{t+1},a; \boldsymbol{\theta}_t^{-})$ is the action values generated by the target network. 
The DQN loss function is given by
\begin{equation} \label{DQN loss}
    J^{\footnotesize(DQN)}_t = (\beta_t^Q-Q(\boldsymbol{s}_t,a_t;\boldsymbol{\theta}_t))^2.
\end{equation} The objective of the deep Q-network is to come up with the optimal policy through the minimization of this loss function. In most applications, it is customary for an agent to take the approach of \emph{exploration and exploitation}~\cite{tokic2010adaptive}.
\begin{figure}[h]
\begin{center}
   \includegraphics[width = 8.5cm, height = 5.8cm]{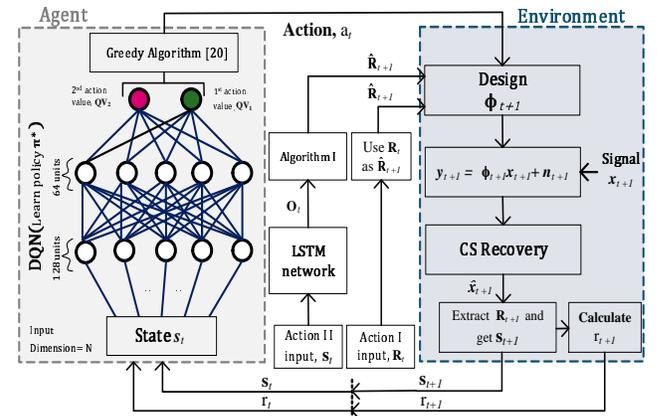}
\end{center}
\vspace{-4.5mm}
   \caption{ \footnotesize{The proposed RL-NCS framework where an agent tries to gain experiences from its interaction with environment and learn the optimal policy $\pi^*$ for designing the measurement matrix.}    
   }
\label{fig:DRL}
\end{figure}
 \vspace*{-\baselineskip} \vspace{0.15cm}
\section{RL-NCS: Reinforcement-Learning-based Non-Uniform CS Framework} \label{CSRL} \vspace*{-\baselineskip} \vspace{0.15cm} In the non-uniform CS technique, it is required to design the measurement matrix in addition to sensing and reconstruction of signals. Fig. \ref{fig:DRL} depicts the way we defined these processes in the environment part of our framework. At each time step, in the agent part (left), a neural network generates action values after receiving input from the environment (right). The measurement matrix $\boldsymbol{\Phi}_{t+1}$ is designed based on the course of action $a_t$ taken by the agent. After CS recovery, the agent receives the new state $\boldsymbol{s}_{t+1}$ and the reward $r_{t+1}$ as feedback. This feedback helps the agent to take the action that results in a higher reward. The goal of the agent is to learn the optimal policy of designing $\boldsymbol{\Phi}$ after a certain number of interactions with the environment. To fully realize the connection between agent and environment, it is necessary to formulate the state space ($\boldsymbol{S}$), action space ($\boldsymbol{A}$), and reward function ($r$). 



\vspace*{-\baselineskip} \vspace{0.15cm}
\subsection{State and Action Space}  Let $\boldsymbol{R}_{t+1}$ be the set of indices of signal coefficients that are in the ROI at time $t+1$. After extracting the $\boldsymbol{R}_{t+1}$\footnote{Depending on the application, the definition of ROI might be different for different scenarios and it can be extracted from the reconstructed signal. For example, ROI can simply be the support of the signal.}, we define $\boldsymbol{s}_{t+1}$, the binary state vector at time $t+1$ by its elements    
\begin{equation} \label{state} \small{
      s_{t+1}^{(l)}  =\left\{
      \begin{array}{@{}ll@{}}
          1 , & \text{if} \hspace{0.1cm} l^{th} \hspace{0.1cm} \text{coef. of signal is in $\boldsymbol{R}_{t+1}$}, ~\\
         0, & \text{otherwise},
      \end{array}\right.}
\end{equation} 
where $s_{t+1}^{(l)}$ is the $l^{th}$ element of $\boldsymbol{s}_{t+1}$ for $1\leq l\leq N$. 


In RL-NCS, the agent needs to decide about the strategy to design the measurement matrix at each time step. For that, we have devised the actions for the action space as following:

\textbf{Action $\mathcal{I}$} (Direct Approach): %
The ROI at time $t$, $\boldsymbol{R}_t$, can be kept unchanged and used readily as the predicted ROI for next step, $\hat{\boldsymbol{R}}_{t+1}$. That is $\hat{\boldsymbol{R}}_{t+1}=\boldsymbol{R}_t$.

\textbf{Action $\mathcal{II}$} (Learning Approach): For rapidly varying signals, the ROI changes relatively fast and the importance levels for most of the coefficients change over time. However, some coefficients still retain the same importance levels. Therefore, a learning mechanism can be developed to identify these two groups of coefficients.
For this, we train an LSTM network, with parameters $\boldsymbol{\Psi}_t$, to extract valuable \emph{correlation information from long sequence of states}.
The output of the LSTM network, $\boldsymbol{O}_t$, helps us to determine $\hat{\boldsymbol{R}}_{t+1}$ using certain confidence bounds. Let $\boldsymbol{I}_t=\{1,2, \ldots, N\}\backslash \boldsymbol{R}_t$ denote the set of indices of estimated non-ROI coefficients at time $t$. We can predict the $\hat{\boldsymbol{R}}_{t+1}$ using \emph{Algorithm \ref{alg:ROI_up}}.

\begin{algorithm}[b]
\caption{Update Rule for ROI}\label{alg:ROI_up}
\algsetup{
linenosize=\footnotesize,
linenodelimiter=:
}
\begin{algorithmic}[1]
\small
\STATE \textbf{Input:} $\hat{\boldsymbol{R}}_{t+1}$ = \{ \}, $\boldsymbol{I}_t$, $\boldsymbol{O}_t$, \hspace{0.05cm} $Th_{up}$ $\in$ [0,1] \hspace{0.05cm} \& \hspace{0.05cm} $Th_{low}$ \hspace{0.05cm} $\in$ [0,1]
\\ 
\STATE $ \textbf{for}\; \text{steps}\; \hspace{0.1cm} j \in \left \{1,2, \ldots N \right \} \; \textbf{do}$
\STATE $\qquad$ $\textbf{if} \hspace{0.1cm} j \in \boldsymbol{R}_t \hspace{0.1cm} \& \hspace{0.1cm} \boldsymbol{O}^{(j)}_t \geq Th_{low}\; \textbf{do}$
\STATE $\qquad$ $\qquad$ $\hat{\boldsymbol{R}}_{t+1} \leftarrow \hat{\boldsymbol{R}}_{t+1} \cup \{j\}$
\\ 
\STATE $\qquad$ $\textbf{else if} \hspace{0.1cm} j \in \boldsymbol{I}_t \hspace{0.1cm}\& \hspace{0.1cm}\boldsymbol{O}^{(j)}_t \geq Th_{up}\; \textbf{do}$
\STATE $\qquad$ $\qquad$ $\hat{\boldsymbol{R}}_{t+1} \leftarrow \hat{\boldsymbol{R}}_{t+1} \cup \{j\}$ \\ $\qquad$ \textbf{end if} 
\\  \textbf{end for}
\STATE \textbf{Output:} $\hat{\boldsymbol{R}}_{t+1}$ (predicted ROI for next time step)
\end{algorithmic}

\end{algorithm}

Here, $\boldsymbol{O}_t^{(j)}$ is the $j^{th}$ element of the LSTM output vector at time $t$. The role of the lower confidence bound, $Th_{low}$, is to select indices in $\boldsymbol{R}_t$ that can be included in $\hat{\boldsymbol{R}}_{t+1}$. On the contrary, the upper confidence bound $Th_{up}$ is used to select indices for $\hat{\boldsymbol{R}}_{t+1}$ from set $\boldsymbol{I}_t$. Since, it is more likely that most of the coefficients retain the same importance levels as before, we set $Th_{up} > Th_{low}$. After forming $\hat{\boldsymbol{R}}_{t+1}$, the measurement matrix can be designed using the same procedure as Action I.
The weighted cross-entropy loss function for the LSTM network is given by
\begin{equation} \label{LSTM loss}
    \small{J^{\footnotesize{(LS)}}_t = -\omega(\boldsymbol{O}^{\small{(tar)}}_t log(\boldsymbol{O}_t))- (\boldsymbol{1}- \boldsymbol{O}^{\small{(tar)}}_t) log(\boldsymbol{1} - \boldsymbol{O}_t) },
\end{equation} where $\boldsymbol{O}^{\small{(tar)}}_t$ works as the target vector for the output of LSTM, $\boldsymbol{O}_t$. A positional weight, $\omega$, is used for the balance between true positive and false negative count.
\subsection{Reward Function} \label{Reward Function} 
We have formulated the reward function in terms of precision and recall defined as
\begin{equation} \label{recall}
        \small{Recall}= \frac{\small{\text{$TP$}}}{|\small{\text{ $\boldsymbol{R}_{t+1}$}}|}, \small{Precision}= \frac{\small{\text{$TP$}}}{|\small{\text{$\hat{\boldsymbol{R}}_{t+1}$}}|},
\end{equation} where $|.|$ represents the cardinality of the set. The variable $TP$ (true positive) indicates the number of elements in the predicted ROI, $\hat{\boldsymbol{R}}_{t+1}$ (predicted by the agent), that falls in the estimated ROI, $\boldsymbol{R}_{t+1}$. It should be noted that the $\boldsymbol{R}_{t+1}$ is extracted from the recovered signal $\hat{x}_{t+1}$ after the CS recovery (as shown in Fig.~\ref{fig:DRL}). Using (\ref{recall}), we calculate the \emph{reward} as
\begin{equation} \label{reward}
    r_{t+1}=\alpha * Precision + (2-\alpha) * Recall.
\end{equation} Here, $\alpha$ indicate the influence of precision and recall on determining the reward at each step. Considering the objective of our work, we set $\alpha<1$. 

\vspace*{-\baselineskip} \vspace{0.1cm}
\subsection{Multi-task Training} \label{training} 
\begin{figure}[htb] 
\begin{center}
   \includegraphics[width=8cm]{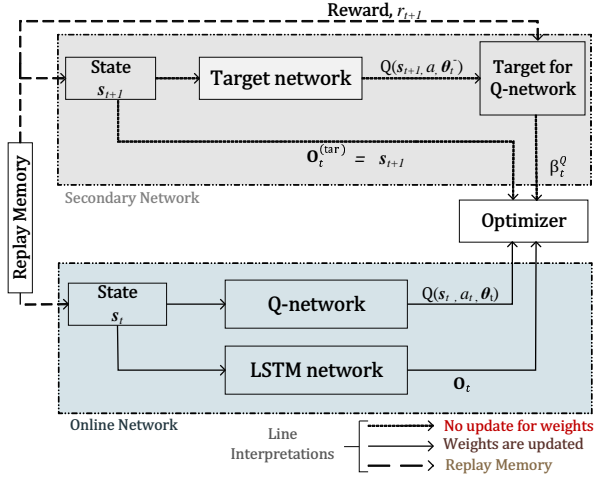}
\end{center}

   \caption{\footnotesize{Training procedure of the online network with proper utilization of replay memory. The purpose of the secondary network is to generate targets for Q-network and LSTM network. 
   }               
   }
\label{fig:LSTM}
\end{figure}

We have used two different neural networks as the online part of our framework as shown in Fig. \ref{fig:LSTM}. One is an LSTM network with 200 hidden units and  the other one is a Q-network that uses two fully connected hidden layers with 128 and 64 neurons, respectively. At each time step $t$, we define the total loss as
\begin{equation} \label{total-loss}
    J_t = (1-\lambda) J_t^{\small{(DQN)}} + \lambda J^{\small{(LS)}}_t,
\end{equation}
where $0 \leq \lambda \leq 1$ is a regularization parameter that controls the flow of LSTM loss,  $J_t^{\small{(LS)}}$, to the optimizer. This way of optimization facilitates the training of LSTM without a dataset. At a certain stage of training, only the Q-network gets tuned until it finds the optimal policy. The complete RL-NCS algorithm is presented in Algorithm \ref{alg:RL-NCS}.

 We have trained Q-network for $T_{max}$ steps and LSTM network for one-third of that steps to have better generalization ability. The episode length ($\tau$) is set to be 100 for the Q-network which also uses a learning rate of 0.05. The learning rate has been reduced to 75$\%$ after each 5000 steps. Furthermore, we have used Rectified Linear Unit (Relu) activation for the hidden layers of the Q-network with the Sigmoid activation at the output. 
 Throughout the training process, the agent (Q-network) follows $\epsilon$-greedy algorithm~\cite{tokic2010adaptive} to take the best action at each step. The value of $\epsilon$ is set to be 1 and decreases to 0 with a decay rate of $\epsilon_{decay}$. The target for Q-network is given by [\ref{target}] while the state $\boldsymbol{s}_{t+1}$ works as the LSTM target. Furthermore, the output of the LSTM is used for Algorithm I to predict the ROI. Here, the LSTM tries to \emph{learn the pattern of changes in consecutive signals} over its course of training. If, in any case, the LSTM fails to learn the pattern, the agent has the option to choose Action I. The Q-network identifies this failure using the rewards obtained from the environment and the range of this reward is between 0 to 2. 
 For high transition probabilities, Action I works as a backup for Action II even though the performance gain might not be similar. 
\begin{algorithm}[htb]
\caption{The RL-NCS Algorithm }\label{alg:RL-NCS}
\algsetup{
linenosize=\footnotesize,
linenodelimiter=:
}
\begin{algorithmic}[1]
\small
\STATE \textbf{Input:}
\hspace{0.15cm} $\boldsymbol{\theta}_t, \hspace{0.1cm} \boldsymbol{\theta}_t^{-} \text{and} \hspace{0.1cm} \boldsymbol{\Psi}_t$ $\colon$ initialize weights (random normal), \\  $\boldsymbol{s}_1$ $\colon$ initialize the state (binary random) 
\\ 
\STATE $ \textbf{for}\; \text{steps}\; t \in \left \{1,2, \ldots T_{max} \right \} \; \textbf{do}$
\STATE $\qquad$ Input $\boldsymbol{s}_t$ to the Q-network to get $a_t$ and then design $\boldsymbol{\Phi}_{t+1}$ 
\STATE $\qquad$ Extract $\boldsymbol{R}_{t+1}$ from $\hat{\boldsymbol{x}}_{t+1}$ to get $\boldsymbol{s}_{t+1}$ and $r_{t+1}$
\STATE $\qquad$ Store the experiences ($\boldsymbol{s}_t$,$a_t$,$\boldsymbol{s}_{t+1}$,$r_{t+1}$) into \\ $\qquad$ \emph{replay memory, $D^{replay}$}.
\STATE $\qquad$ Sample a mini-batch of $Z$ transitions, SE, from $D^{replay}$
\\
\STATE $\qquad$ \textbf{for} each sample e=($\boldsymbol{s}_t$,$a_t$,$\boldsymbol{s}_{t+1}$,$r_{t+1}$) in SE \textbf{do} \\ $\qquad$ \hspace{0.195cm} $\boldsymbol{O}^{(tar)}_t$ = $\boldsymbol{s}_{t+1}$\\  $\qquad$ \hspace{0.22cm} $ \beta_{t}^{Q} = r_{t+1} + \gamma \max\limits_{a \in A} \boldsymbol{Q}(\boldsymbol{s}_{t+1},a; \boldsymbol{\theta}_t^{-}) $\\ $\qquad$ \textbf{end for} 
\STATE $\qquad$ Calculate the loss using (\ref{total-loss}) and decrease $\lambda $ by $\Delta \lambda$

\STATE $\qquad$ Update $\boldsymbol{\Psi}_t$ and $\boldsymbol{\theta}_t$ using gradient descent method 
\STATE $\qquad$ \textbf{if} \; t mod $\tau$ = 0; then $\boldsymbol{\theta}_t^{-}$  $\longleftarrow$ $\boldsymbol{\theta}_t$ \textbf{end if} 
\STATE $\qquad$ $\boldsymbol{s}_t$ $\longleftarrow$ $\boldsymbol{s}_{t+1}$ 
\\  \textbf{end for}
\STATE \textbf{Output:} An optimal policy for designing $\boldsymbol{\Phi}$
\end{algorithmic}
\end{algorithm}

 

\vspace*{-\baselineskip} \vspace{0.15cm}
\section{Numerical Experiments} \label{NE} \vspace*{-\baselineskip} \vspace{0.15cm}
In this section, a detailed description of the executed experiments are presented along with the outcome of the experiments. To underscore the effectiveness of our method, the primary performance metric that has been used here is
\begin{equation} \label{TNMSE}
    \text{TNMSE} =\frac{1}{\tau}\sum_{t=1}^\tau \frac{||\boldsymbol{x}_t-\hat{\boldsymbol{x}_t||_2^2}}{||\boldsymbol{x}_t||_2^2}.
\end{equation}
TNMSE stands for time averaged normalized mean square error over the time period $\tau$ and $\hat{\boldsymbol{x}}_t$ is the reconstructed signal while $\boldsymbol{x}_t$ is its original counterpart. $||.||_2$ denotes the $\ell_2$-norm of a vector.  

 At each step, the estimation of the signal is obtained using the $\ell_1$ minimization recovery algorithm given by
 \vspace{0.15cm}
\begin{equation} \label{minimization}
    \hat{\boldsymbol{x}}_t = \text{arg min} {||\boldsymbol{x}_t||}_1, \hspace{0.2cm} s.t. {||\boldsymbol{y}_t- \boldsymbol{\Phi}_t\boldsymbol{x}_t||_2} \leq \mu ,
\end{equation}  \vspace{0.15cm} where $\hat{\boldsymbol{x}}_t$ is the estimated signal determined through the minimization of $||\boldsymbol{x}||_1$ = $\sum_{n} |x^{(n)}|$. The error bound $\mu$ is assigned to have a value of $\sigma_{n}\sqrt{M}$.

After predicting the ROI, we distribute the sensing energy to the columns of $\boldsymbol{\Phi}$ according to the prediction. We set the $\ell_2$-norm of the $n^{th}$ column as
\vspace*{-\baselineskip} \vspace{0.15cm}
\begin{equation} \label{Sensing energy}
        e^{(n)} =\small{ \sqrt{N} \hspace{0.1cm}\frac{{\eta^{(n)}}}{\sqrt{\sum_{n} {\{\eta^{(n)}\}}^2}}},
\end{equation} 
 where $\eta^{(n)}$ is the importance level for $n^{th}$ coefficient and $N$ represents the total number of coefficients. 
The distribution of the energy is done in a way such that the columns corresponding to ROI coefficients receive more energy than the non-ROI counterpart and the criteria of energy constraint, $||\boldsymbol{\Phi}||_F^2 =N$, is met.
On the other hand, for uniform sampling, the columns of the matrices are scaled to have unit norm.

\vspace*{-\baselineskip} \vspace{0.15cm}
\subsection{Simulations with sparse signals in the canonical basis} \label{Sparse domain}
At first, signals that are sparse in the canonical basis are employed for all simulations. Therefore, the ROI is considered to be the support of the signal. Furthermore, it has been assumed that the signals in the time series are correlated. 
To establish the correlation in value, the evolution of the value of $n^{th}$ signal coefficient $w^{(n)}_t$ can be modelled as $(1-\rho)w^{(n)}_{t-1} + \rho v^{(n)}_t$. Here, the correlation parameter $\rho$ dictates the degree of correlation and assumes a value between 0 and 1. 
The term $v^{(n)}_t$ is modelled with zero mean Gaussian distribution, $\mathcal{N}(0,\sigma_L^2)$, for imposing variations among two consecutive time steps. 
For describing the ROI, we consider a binary vector $\boldsymbol{d}_t = [ d^{(1)}_t, \ldots ,d^{(N)}_t ]^T$. A value of 1 for $d_t^{(n)}$ indicates $n^{th}$ coefficient is in the ROI and a zero value indicates otherwise. All the values of $\boldsymbol{d}_t$ are assumed to be independent of each other.

In order to establish the correlation among the ROI of two consecutive time steps, a Markov chain process is defined for each signal coefficient. That is, the modelling of transition probabilities, $tp_{01} = \mathrm{P}\{d^{(n)}_t=1|d^{(n)}_{t-1} = 0\}$ and $tp_{10} = \kappa *tp_{01}/ (1-\kappa)$, is achieved by Markov chain process. The term $\kappa$ denotes the sparsity level and can be expressed as $\kappa = \mathrm{P}\{d^{(n)}_t =1\}$. Finally, the coefficients of the signal at each time step can be formulated as $x^{(n)}_t = w^{(n)}_t d^{(n)}_t + b^{(n)}_t(1-d^{(n)}_t)$. Small (or non-ROI) coefficients $b^{(n)}_t$ are modelled with $\mathcal{N}(0,\sigma_S^2)$. The simulation parameters for the CS environment are set as follows. 
The correlation parameter is set as $\rho = 0.2$. Standard deviation(SD) of large and small coefficients are $\sigma_L = 5$ and $\sigma_S = 0.01$, respectively. Furthermore, the value of $\eta^{(n)}$ is 0.7 for ROI coefficients and 0.3 for non-ROI coefficients. All other simulation parameters are $\gamma = 0.1$, $\alpha = 0.5$, $\omega = 5 $, $Th_{up}=0.8$, $Th_{low} =0.1$, $\tau = 100$, $\lambda = 1$, $\Delta \lambda = 1/10000$, $\epsilon_{decay} = 1/10000$ and $T_{max} = 30000$, $N=200$.  

\begin{figure}[htb] 
\begin{center}
   \includegraphics[width=8.7cm]{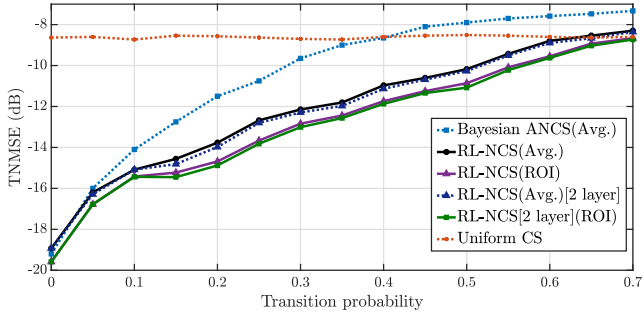}
\end{center}
\vspace{-4mm}
   \caption{\footnotesize{Performance evaluation of $\textit{l}_1$ recovery and proposed sampling method (RL-NCS) for different values of $tp_{01}$. 
   For setup: $N=200$, $M = 60$, SNR = 20dB, and total sensing energy = $N$.}}
\label{fig:transition}

\end{figure}

\begin{figure}[htb] 
\begin{center}
   \includegraphics[width=8.7cm]{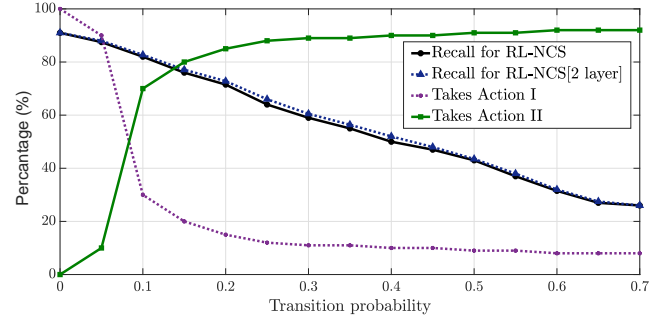}
\end{center}
\vspace{-4mm}
   \caption{ \footnotesize{The percentage of recall and percentage of chosen type of action (by agent) for different transition probability, $tp_{01}$. 
   } 
   }
\label{fig:recall}
\end{figure}

For performance evaluation, TNMSE (dB) for different values of transition probabilities is shown in Fig. \ref{fig:transition}. We have averaged the reconstruction error over the whole episode. It can be observed that the proposed sampling method outperforms the uniform CS upto $tp_{01} \leq 0.62$. Unlike Bayesian ANCS~\cite{zaeemzadeh2017adaptive}, RL-NCS has significant performance gains over uniform CS for high transition probabilities. We have also shown the TNMSE only for the coefficients in ROI and it is observable that ROI coefficients are reconstructed with more accuracy compared to Non-ROI counterpart. A slight improvement is noticed in the performance gain for an extra LSTM layer. In order to comprehend this improvement, the percentage value of recall for different $tp_{01}$ is shown in Fig. \ref{fig:recall}. We can see that two-layered LSTM helps the agent to infer the next ROI more accurately. Furthermore, it is also shown that the trend of choosing second action increases for signals with rapid variations. 

\begin{figure}[htb]
\begin{center}
   \includegraphics[width=8.7cm]{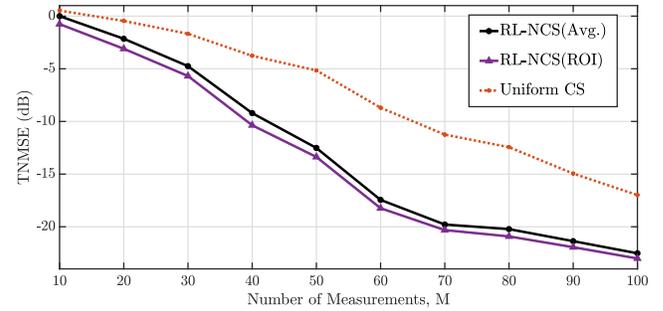}
\end{center}
\vspace{-4mm}
   \caption{\footnotesize{ Time averaged recovery error (in dB) for different number of measurements. 
   For setup: $N=200$, $tp_{01}$ = 0.02, Input SNR = 20dB.}
   }
\label{fig:Measurement}
\end{figure}

\begin{figure}[htb]
\begin{center}
   \includegraphics[width=8.7cm]{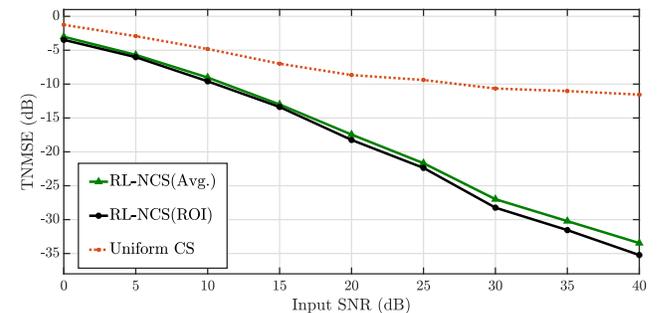}
\end{center}
\vspace{-4mm}
   \caption{\footnotesize{ Performance evaluation of recovery algorithm for different values of input SNR.
   For setup: $N=200$, $M = 60$, $tp_{01}$ = 0.02 and $T= 30$.}
   }
\label{fig:in_SNR}
\end{figure}

Fig. \ref{fig:Measurement} shows the recovery performance of uniform and non-uniform CS for different number of measurements. It can be seen that a performance gain up to 8.75 dB (for M = 60) is obtainable by employing RL-NCS. In addition, the reconstruction performance of $\textit{l}_1$ recovery algorithm boosts up even with less number of measurements. The difference in performance gain is easily discernible. For example, uniform sampling requires 63$\%$ more measurements than RL-NCS to achieve a -15 dB gain. Furthermore, the noise associated with the sampling step usually increases the reconstruction error. Our proposed method is effective even in the regime of low SNR. Fig. \ref{fig:in_SNR} depicts the performance gain achieved by our method over other sampling method for different input SNRs. 

\vspace*{-\baselineskip} \vspace{0.15cm}
\subsection{Simulations with sparse signals in the DCT domain} \label{DCT domain} 
We have also carried out experiments with signals that are sparse in the DCT domain. Let, $\boldsymbol{z}_t = \boldsymbol{\Theta} \boldsymbol{x}_t$ be the sparse representation of the signal $\boldsymbol{x}_t$, obtained by the DCT transform matrix $\boldsymbol{\Theta}$. To establish the correlation between signals, same procedure explained in Section~\ref{Sparse domain} has been followed. Since ROI and support of the signal are not the same in this case, a new set of binary Markov processes is used to model the variation of ROI. This Markov process uses the same parameters $\kappa$ and $tp_{01}$ to impose a certain degree of randomness along with the correlation. Therefore, we have a setup where support of $\boldsymbol{z}_t$ and ROI in $\boldsymbol{x}_t$ change at the same rate. To reconstruct $\hat{\boldsymbol{z}}_t$ we employ   
\begin{equation} \label{DCT_minimization}
    \hat{\boldsymbol{z}}_t = \text{arg min} {||\boldsymbol{z}_t||}_1, \hspace{0.2cm} s.t. {||\boldsymbol{y}_t- \boldsymbol{\Phi}_t\boldsymbol{\Theta}^T\boldsymbol{z}_t||_2} \leq \mu
\end{equation} that eventually leads us to $\boldsymbol{x}_t = \boldsymbol{\Theta}^T \boldsymbol{z}_t$. 

\begin{figure}[htb]
\begin{center}
   \includegraphics[width=8.7cm]{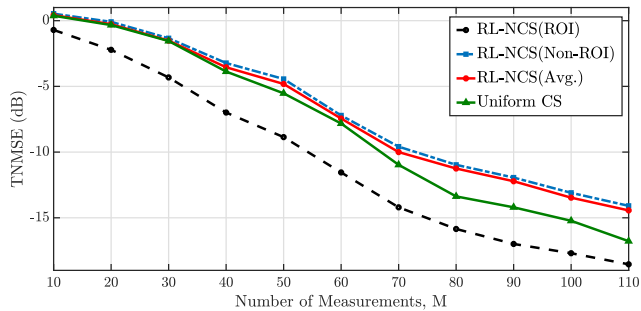}
\end{center}
\vspace{-4mm}
   \caption{\footnotesize{ TMSNE error (in dB) for different number of measurements (M). For setup: $N =200$, $T = 30$, SNR = 20 dB and fault-rate = 0.10.}
   }
\label{fig:DCT_meas}
\end{figure}

\begin{figure}[htb]
\begin{center}
   \includegraphics[width=8.7cm]{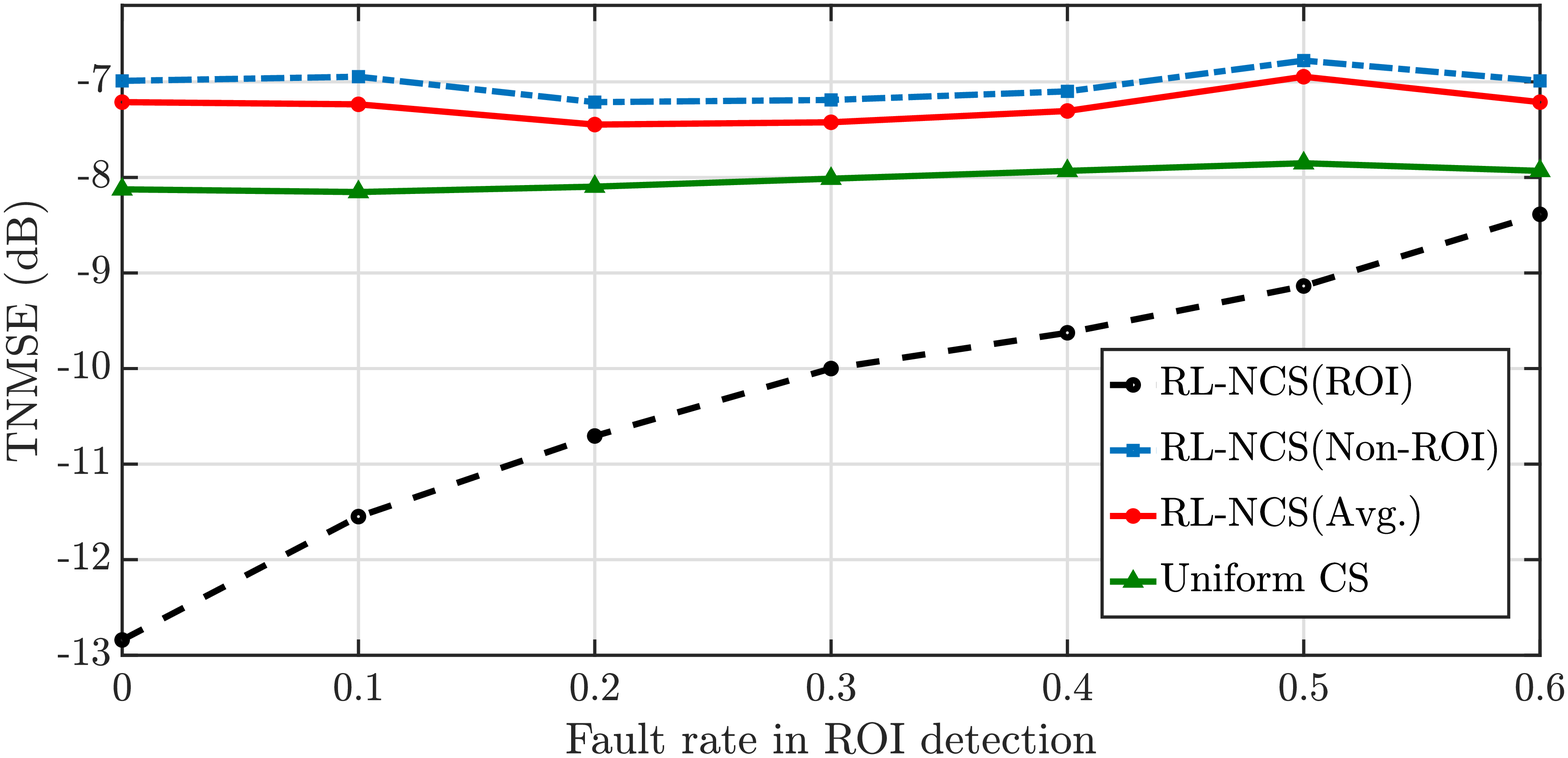}
\end{center}
\vspace{-4mm}
   \caption{\footnotesize{ TMSNE error (in dB) for different fault-rate ($\%$) in ROI detection. 
   }
   }
\label{fig:fault_rate}
\end{figure}

Depending on the applications, there are different methods for figuring out the ROI of signal. Furthermore, we also assume that the ROI detection method may give faulty observations. The performance of RL-NCS for different number of measurements are depicted in Fig. \ref{fig:DCT_meas}. Considering only the ROI coefficients, the RL-NCS algorithm has a significant performance gain over uniform CS even though it loses out on total reconstruction error. In Fig. \ref{fig:fault_rate}, the performance for non-uniform CS is shown for different fault rates in ROI detection. As the estimation of ROI gets more erroneous, it gets harder for the agent to fix the location of ROI. However, even for 60$\%$ fault rate, RL-NCS performs better than uniform CS when it comes to the  reconstruction of ROI coefficients.              

\vspace*{-\baselineskip} 
 \section{Conclusion} \label{Discussion} \vspace*{-\baselineskip} \vspace{0.15cm}
 We proposed a framework with a view to achieving non-uniform compressed sensing for time-varying signals. To overcome the limitation of adaptation with fast varying signals, an LSTM network along with a manually tailored action is introduced.
 A deep Q-network approach has also been employed here whose purpose is to choose the best action for designing measurement matrix at each time step. With the joint training method, we introduced an elegant solution for training an LSTM-based learning mechanism while obtaining an optimal policy from Q-network in the end. Moreover, we have carried out experiments for sparse signals both in the canonical basis and in the DCT domain. The results presented here show that our method achieves significant performance gain over uniform CS. 

\vspace*{-\baselineskip} \vspace{0.1cm}
\bibliographystyle{IEEEbib}
 
\bibliographystyle{IEEEbib}

\begin{thebibliography}{00}
\footnotesize{
\bibitem{candes2008introduction}
Candes EJ, Wakin MB. An introduction to compressive sampling [a sensing/sampling paradigm that goes against the common knowledge in data acquisition]. IEEE signal processing magazine. 2008;25(2):21-30.
\vspace{-3mm}

\bibitem{donoho2006compressed}
Donoho DL. Compressed sensing. IEEE Transactions on information theory. 2006 Apr 1;52(4):1289-306.
\vspace{-3mm}

\bibitem{vaswani2008kalman}
Vaswani N. Kalman filtered compressed sensing. In2008 15th IEEE International Conference on Image Processing 2008 Oct 12 (pp. 893-896). IEEE.
\vspace{-3mm}

\bibitem{asif2010streaming}
Asif MS, Reddy D, Boufounos PT, Veeraraghavan A. Streaming compressive sensing for high-speed periodic videos. In2010 IEEE International Conference on Image Processing 2010 Sep 26 (pp. 3373-3376). IEEE.
\vspace{-3mm}

\bibitem{zaeemzadeh2018cospot}
Zaeemzadeh A, Joneidi M, Rahnavard N, Qi GJ. Co-SpOT: Cooperative spectrum opportunity detection using bayesian clustering in spectrum-heterogeneous cognitive radio networks. IEEE Transactions on Cognitive Communications and Networking. 2017 Dec 27;4(2):206-19.
\vspace{-3mm}

\bibitem{Shahrasbi2016HMT}
Shahrasbi B, Rahnavard N. Model-based nonuniform compressive sampling and recovery of natural images utilizing a wavelet-domain universal hidden Markov model. IEEE Transactions on Signal Processing. 2016 Sep 29;65(1):95-104.
\vspace{-3mm}

\bibitem{Rahnavard2011}
Rahnavard N, Talari A, Shahrasbi B. Non-uniform compressive sensing. In2011 49th Annual Allerton Conference on Communication, Control, and Computing (Allerton) 2011 Sep 28 (pp. 212-219). IEEE.
\vspace{-5.5mm}

\bibitem{mnih2013playing}
Mnih V, Kavukcuoglu K, Silver D, Graves A, Antonoglou I, Wierstra D, Riedmiller M. Playing atari with deep reinforcement learning. arXiv preprint arXiv:1312.5602. 2013 Dec 19.
\vspace{-3mm}

\bibitem{chincoli2018self}
Chincoli M, Liotta A. Self-learning power control in wireless sensor networks. Sensors. 2018 Feb;18(2):375.
\vspace{-3mm}

\bibitem{le2016reinforcement}
Le TT, Moh S. Reinforcement-Learning-Based Topology Control for Wireless Sensor Networks. Proceedings of the Grid and Distributed Computing 2016. 2016:22-7.
\vspace{-3mm}

\bibitem{malloy2014near}
Malloy ML, Nowak RD. Near-optimal adaptive compressed sensing. IEEE Transactions on Information Theory. 2014 Jul;60(7):4001-12.
\vspace{-3mm}

\bibitem{braun2015info}
Braun G, Pokutta S, Xie Y. Info-greedy sequential adaptive compressed sensing. IEEE Journal of selected topics in signal processing. 2015 Jun;9(4):601-11.
\vspace{-3mm}

\bibitem{ziniel2013dynamic}
Ziniel J, Schniter P. Dynamic compressive sensing of time-varying signals via approximate message passing. IEEE transactions on signal processing. 2013 Nov 1;61(21):5270-84.
\vspace{-3mm}

\bibitem{Ziniel2014}
Shahrasbi B, Talari A, Rahnavard N. TC-CSBP: Compressive sensing for time-correlated data based on belief propagation. In2011 45th Annual Conference on Information Sciences and Systems 2011 Mar 23 (pp. 1-6). IEEE.
\vspace{-3mm}

\bibitem{vaswani2010ls}
Vaswani N. LS-CS-residual (LS-CS): compressive sensing on least squares residual. IEEE Transactions on Signal Processing. 2010 Aug;58(8):4108-20.
\vspace{-3mm}

\bibitem{zaeemzadeh2017adaptive}
Zaeemzadeh A, Joneidi M, Rahnavard N. Adaptive non-uniform compressive sampling for time-varying signals. In2017 51st Annual Conference on Information Sciences and Systems (CISS) 2017 Mar 22 (pp. 1-6). IEEE.
\vspace{-3mm}


\bibitem{greff2016lstm}
Greff K, Srivastava RK, Koutník J, Steunebrink BR, Schmidhuber J. LSTM: A search space odyssey. IEEE transactions on neural networks and learning systems. 2016 Jul 8;28(10):2222-32.
\vspace{-3mm}

\bibitem{watkins1992q}
Watkins CJ, Dayan P. Q-learning. Machine learning. 1992 May 1;8(3-4):279-92.
\vspace{-3mm}


\bibitem{mnih2015human}
Mnih V, Kavukcuoglu K, Silver D, Rusu AA, Veness J, Bellemare MG, Graves A, Riedmiller M, Fidjeland AK, Ostrovski G, Petersen S. Human-level control through deep reinforcement learning. Nature. 2015 Feb;518(7540):529.
\vspace{-3mm}

\bibitem{tokic2010adaptive}
Tokic M. Adaptive e-greedy exploration in reinforcement learning based on value differences. InAnnual Conference on Artificial Intelligence 2010 Sep 21 (pp. 203-210). Springer, Berlin, Heidelberg.
\vspace{-3mm}


}
\end{thebibliography}
\end{document}